\documentclass[10pt,twocolumn,letterpaper]{article}

\usepackage{iccv}
\usepackage{times}
\usepackage{epsfig}
\usepackage{graphicx}
\usepackage{amsmath}
\usepackage{amssymb}
\usepackage{booktabs}       
\usepackage{multirow}
\usepackage{multicol}
\usepackage{algorithm}
\usepackage{algpseudocode}

\usepackage[breaklinks=true,bookmarks=false]{hyperref}  

\iccvfinalcopy 

\def\iccvPaperID{6094} 

\ificcvfinal\fi

\begin{document}

\title{Inversion-by-Inversion: Exemplar-based Sketch-to-Photo Synthesis via Stochastic Differential Equations without Training}

\author{Ximing Xing\\
Beihang University\\
School of Software\\
{\tt\small ximingxing@buaa.edu.cn}
\and
Chuang Wang\\
Beihang University\\
School of Software\\
{\tt\small 18377011@buaa.edu.cn}
\and
Haitao Zhou\\
Beihang University\\
School of Software\\
{\tt\small 18377221@buaa.edu.cn}
\and
Zhihao Hu\\
Beihang University\\
School of Software\\
{\tt\small huzhihao@buaa.edu.cn}
\and
Chongxuan Li\\
Renmin University of China\\
The Gaoling School of AI\\
{\tt\small chongxuanli@ruc.edu.cn}
\and
Dong Xu\\
The University of Hong Kong\\
The Department of Computer Science\\
{\tt\small dongxu@cs.hku.hk}
\and
Qian Yu\\
Beihang University\\
School of Software\\
{\tt\small qianyu@buaa.edu.cn}
}



\maketitle
\ificcvfinal\thispagestyle{empty}\fi

\begin{abstract}
\vspace{-2mm}
Exemplar-based sketch-to-photo synthesis allows users to generate photo-realistic images based on sketches. Recently, diffusion-based methods have achieved impressive performance on image generation tasks, enabling highly-flexible control through text-driven generation or energy functions. However, generating photo-realistic images with color and texture from sketch images remains challenging for diffusion models. Sketches typically consist of only a few strokes, with most regions left blank,  making it difficult for diffusion-based methods to produce photo-realistic images. 
In this work, we propose a two-stage method named ``Inversion-by-Inversion" for exemplar-based sketch-to-photo synthesis. This approach includes shape-enhancing inversion and full-control inversion. During the shape-enhancing inversion process,  an uncolored photo is generated with the guidance of a shape-energy function. This step is essential to ensure control over the shape of the generated photo. In the full-control inversion process, we propose an appearance-energy function to control the color and texture of the final generated photo.
Importantly, our Inversion-by-Inversion pipeline is training-free and can accept different types of exemplars for color and texture control. We conducted extensive experiments to evaluate our proposed method, and the results demonstrate its effectiveness.
\end{abstract}

\vspace{-4mm}
\section{Introduction}
Sketch is an intuitive and powerful tool for humans to express ideas~\cite{10098211}. With just a few lines, a sketch can convey a complex concept, and humans can easily envision a realistic photo based on a sketch. However, this presents a significant challenge for machines due to the stark contrast between black-and-white sketches and colorful photos. In recent years, the task of sketch-to-photo synthesis has gained increasing attention~\cite{6196209,7335623,10061434,10082973,chen2018sketchygan,chen2020deepfacedrawing,liu2020unsupervised,sketchgan_Wang_2021,xiang2022adversarial}, particularly with the popularity of AI-generated content (AIGC). 

Recently, Stochastic Differential Equations (SDE)~\cite{EestGrad_song_2019, ddpm_ho_2020, scorebased_song_2021,ADM_dhariwal_2021} or diffusion models have recently shown great potential in image generation tasks and have become a driving force in the development of AIGC. Energy-based SDE~\cite{egsde_zhao_2022} has also achieved promising results in image-to-image translation tasks. In \cite{sdedit_meng_2022}, simply applying noise on the input stroke image, the model can progressively remove noise from each pixel and generate a photo-realistic image that is roughly spatial-aligned with the stroke image. EGSDE~\cite{egsde_zhao_2022} further improves generation quality by introducing energy functions to balance faithfulness and realism. Despite the effectiveness of the diffusion models on the aforementioned tasks, they are seldom studied for sketch-based photo synthesis.

Sketches are simple but challenging for diffusion models to handle. Compared to an RGB photo, a sketch lacks colors and visual details, primarily consisting of black pixels with the remaining areas being white. However, the inverse process of SDE, i.e., denoising, is performed in the pixel space. Therefore, it is difficult for SDE to directly generate colorful photo-realistic images from a sketch. An exemplar image can be adapted to provide the missing information of a sketch for synthesis. Nevertheless, a conflict arises when both the sketch and exemplar photos exist during the denoising process. The exemplar photo will suppress the impact of the sketch, thus preventing the generated photo from preserving the geometry structure of the input sketch. 

\begin{figure}[t]
\centering
\includegraphics[width=1\linewidth]{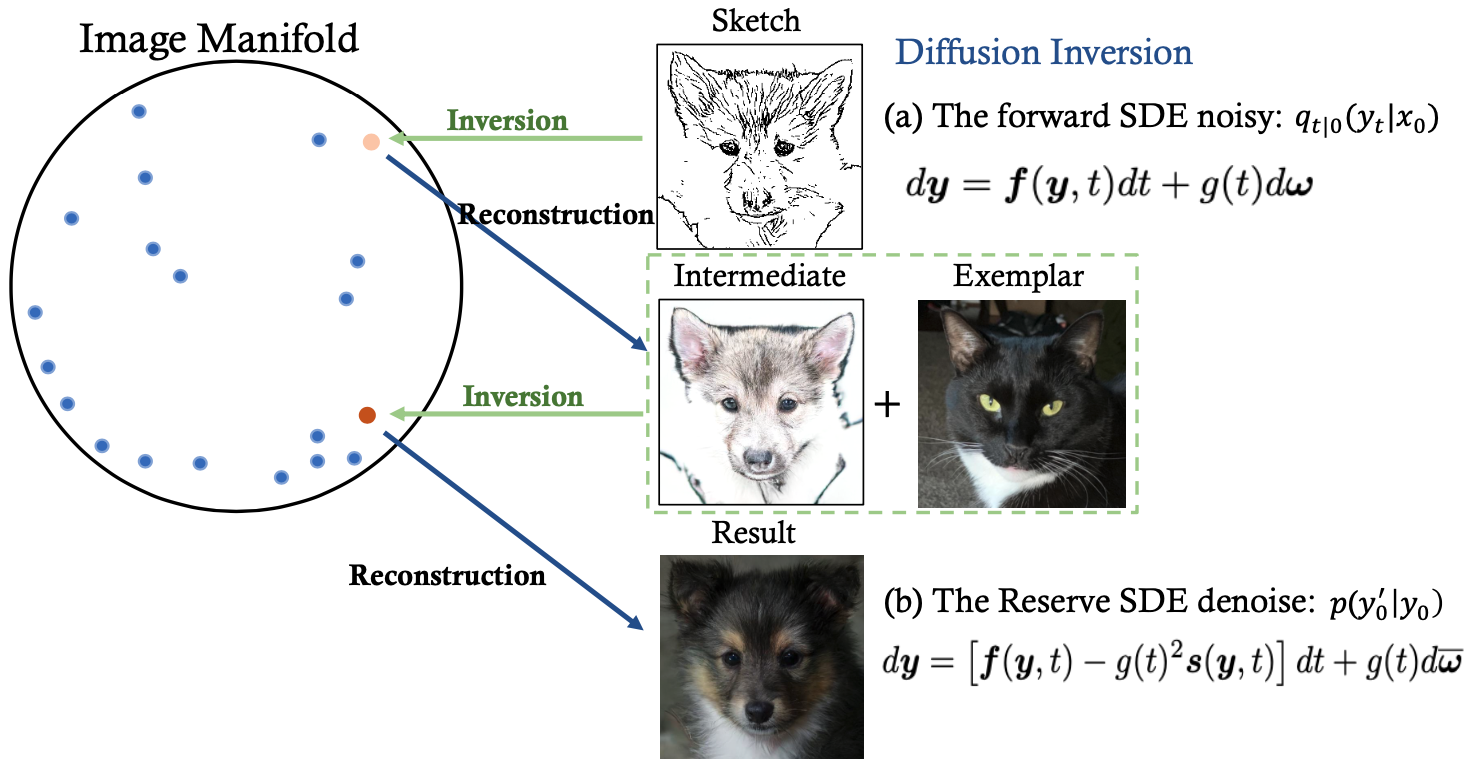}
\caption{
The Illustration of our Inversion-by-Inversion Translation via SDE. 
A large domain gap exists between sketches and the photos, making it hard to directly generate the final output photo from a sketch. Therefore, we propose an Inversion-by-Inversion sketch-to-image translation method, which involves two stages of inversion, namely the shape-enhancing inversion and the full-control inversion. In our Inversion-by-Inversion method, we first generate  intermediate uncolored photos from sketches using our shape-enhancing inversion and then generate the final results using our full-control inversion.
}
\label{fig:tears}
\end{figure}

To address the aforementioned issues, we propose a new exemplar-based sketch-to-photo synthesis method called \textit{Inversion-by-Inversion}. Our method consists of two inversion SDE, a shape-enhancing inversion and a full-control inversion. The motivation behind our approach is shown in Figure~\ref{fig:tears}, and an overview of our method is presented in Figure~\ref{fig:overview}.
In our proposed model, a sketch controls the shape of the target photo, while an exemplar image controls other visual features such as colors and textures. Inspired by energy-guided SDE~\cite{egsde_zhao_2022} which uses energy functions to achieve a balance between faithfulness and realism, we design geometry-energy and appearance-energy functions for shape and appearance control respectively, and use them as guidance in the full-control inversion process.
However, we have noticed that the geometric structure of sketches is difficult to preserve in the generated photos, as the shape information provided by the sketches is likely to be lost in the presence the exemplar images. 
Therefore, we introduce a shape-enhancing inversion process before the full-control inversion to strengthen the shape control of the sketches. 
This step is critical to the success of the task, as we will show in the experiment section. 

In the shape-enhancing step, the uniform noise is added to the input sketch, and a geometry-energy function is used to guide the inverse process of SDE, resulting in a uncolored photo that matches the shape of the sketch. In the full-control step, the uniform noise is added to the generated uncolored photo, and both geometry-energy and appearance-energy functions are adopted to guide the inversion of SDE. During this process, the color and detailed texture of the exemplar are gradually added to the uncolored image, while its shape is preserved. After these two steps, we obtain a photo whose shape follows the input sketch while its visual details follow the exemplar image.

It is worth noting that our newly proposed approach \textit{does not require any additional task-specific training steps} and can be easily applied to any other pre-trained SDE models for sketch-to-photo synthesis. 
As shown in Figure~\ref{fig:all_in_one}, the proposed model can accept different types of exemplar images for color and texture control, including photos, stroke images, segmentation maps, or style images. It is also capable of handling free-hand sketches, achieving a balance between shape faithfulness and visual realism. We also conduct in-depth analysis and provide insights to explain how the two-step inversion pipeline benefits the transition from a sketch to an RGB photo.

Our contributions are summarized as follows:
\begin{itemize}
    \item We propose the first SDE-based model named \textit{Inversion-by-Inversion} for exemplar-based sketch-to-photo synthesis, which can synthesize a photo with the shape of the input and the appearance of the exemplar image. 
    \item This new approach is specifically designed for SDE-based models to handle sketch images. With the help of shape-enhancing inversion, the generated photo can preserve the shape of the input sketch. To the best of our knowledge, this is the first SDE-based model that can achieve such a goal using only images. 
    \item We present a shape-energy function and an appearance-energy function for sketch and exemplar control, which are used in the full-control inversion process. 
    \item Extensive experiments are conducted to demonstrate the effectiveness of the proposed method. It significantly outperforms baseline models in terms of visual quality and shape consistency. 
\end{itemize}

\section{Related Work}

\subsection{Sketch-to-Photo Synthesis}
Many sketch-to-photo synthesis methods have been developed in the past few years, which aim to generate a photo-realistic image based on a given input sketch. This task is non-trivial, most previous methods focus on face generation~\cite{4453838,6196209,7335623,10061434,10082973,chen2020deepfacedrawing,li2020deepfacepencil,cheng_2023_adaptively,voynov_2022_sketchMLP,Wang_2022_DiffSketching,controlnet_zhang_2023} or object-level image generation~\cite{chen2018sketchygan,ghosh2019interactive,liu2020unsupervised,xiang2022adversarial,an2023sketchinverter}. SketchyGAN~\cite{chen2018sketchygan} is the first learning-based free-hand sketch-to-photo synthesis method, which is trained on the sketch-photo pairs with class labels. \cite{ghosh2019interactive} makes it possible for users to interact with the sparse sketch and perform the sketch-to-photo generation process in a feedback loop. Inspired by CycleGAN~\cite{zhu2017unpaired}, Xiang~\textit{et.al.}~\cite{xiang2022adversarial} and Liu~\textit{et.al}~\cite{liu2020unsupervised} proposed new sketch-to-photo paradigms based on the cycle consistency. Recently, Wang~\textit{et.al}~\cite{sketchgan_Wang_2021} fine-tuned the pre-trained StyleGAN~\cite{stylegan_Karras_2019} for sketch-faithful images generation. 
Sketch-guided face synthesis is the research direction of Sketch-to-Photo synthesis~\cite{4453838,6196209,7335623,10061434,10082973,chen2020deepfacedrawing}.
Peng~\textit{et.al}~\cite{7335623} employs a two-stage synthesis process to learn face structures through image segmentation.
Most existing sketch-to-photo methods are based on generative adversarial networks (GANs), and their performance heavily depends on the performance of the generation model.
Recently, the diffusion model~\cite{ddpm_ho_2020,stable_diffusion_Rombach_2022} achieved impressive image generation performance. Therefore, developing a new sketch-to-photo synthesis algorithm is desirable based on the effective diffusion model.
Previous research~\cite{cheng_2023_adaptively,voynov_2022_sketchMLP,Wang_2022_DiffSketching,controlnet_zhang_2023} has explored the retraining of diffusion models~\cite{cheng_2023_adaptively,Wang_2022_DiffSketching} or the addition of extra hypernetworks~\cite{voynov_2022_sketchMLP,controlnet_zhang_2023} to incorporate sketches as additional control conditions. However, these approaches require additional training costs to adapt for a single style of sketch. In contrast, the diffusion model framework proposed in this paper is training-free and can adapt to different sketch conditions by adjusting the energy function.

\subsection{Score-based Diffusion Models}

Score-based diffusion models (SBDMs)~\cite{EestGrad_song_2019, ddpm_ho_2020, scorebased_song_2021,ADM_dhariwal_2021} perturb data during the forward process and recover the data in the inversion process. Let $q \left(\boldsymbol{y}_0\right)$ be the unknown data distribution on $\mathbb{R}^{D}$. The forward diffusion process of $\left\{ \boldsymbol{y}_t \right\}_{t \in [0,T]}$ at time $t$ can be represented by the following forward SDE: 
\begin{equation}
d \mathbf{y} = \mathbf{f} (\boldsymbol{y}, t) dt + g(t) d\mathbf{\omega}
\label{eq:foward_sde}
\end{equation}
\noindent in which $\mathbf{f} \left(\cdot, t \right):\mathbb{R}^{D}\xrightarrow{}\mathbb{R}^{D}$ is the drift coefficient, $g(t) \in \mathbb{R}$ is the diffusion coefficient and  $\mathbf{\omega} \in \mathbb{R}^{D}$ is a standard Wiener process. $f(\mathbf{y},t)$ and $g(t)$ are correlated to the noise size and determine the perturbation kernel $q_{t|0} \left(\mathbf{y}_t | \mathbf{y}_0 \right)$ from time 0 to $t$. Note $f(\mathbf{y}, t)$ is usually affine and can be sampled in one step. 

Let $q_t(\boldsymbol{y})$ be the marginal distribution of the SDE at time $t$ in Eq.~\ref{eq:foward_sde}. Its time reversal can be described by another SDE~\cite{scorebased_song_2021}:
\begin{equation}
d\mathbf{y} = \left[\mathbf{f}(\mathbf{y},t) - g(t)^2 \mathbf{s}(\mathbf{y},t)\right]dt + g(t) d\mathbf{\overline{\omega}}
\label{eq:reverse_sde}
\end{equation}
\noindent where $\boldsymbol{\overline{\omega}}$ is a reverse-time standard Wiener process, and $dt$ is an infinitesimal negative timestep. Song \textit{et al.}~\cite{scorebased_song_2021} adopts a score-based model $\boldsymbol{s}(\boldsymbol{y},t)$ to approximate the unknown $\nabla_{\boldsymbol{y}} \log q_t(\boldsymbol{y})$ by score matching~\cite{score_matching_Aapo_2005}, thus inducing a score-based diffusion model (SBDM).

\subsection{Generation Model Inversion}
Generative adversarial network (GAN) inversion techniques~\cite{9953153,9953190,9656731,10044117,9126831,9915620,GAN_inversion_survey} have been explored for real image editing. It projects an image to the latent space of a pre-trained GAN generator, edits the latent code, and finally re-generates an image based on the new latent code.
GAN inversion makes the controllable directions found in latent spaces of the existing trained GANs applicable to editing real images~\cite{stylegan_Karras_2019, IDganInver_zhu_2020},
However, with a low bitrate latent code, previous works have difficulties in preserving high-fidelity details in reconstructed and edited images~\cite{HFganInver_Wang_2022}.
Increasing the size of a latent code can improve the fidelity of GAN inversion but at the cost of inferior editability.

Recently, several studies~\cite{ILVR_Choi_2021,lowdensity_Sehwag_2022,sdedit_meng_2022,edict_wallace_2022,egsde_zhao_2022,equivariant_bao_2023} leveraged score-based diffusion models for image translation due to their powerful generative ability and achieved good results.
Finding an initial noise vector that produces an input image when fed into the diffusion process (known as inversion) is an important problem in denoising diffusion models, with applications for real image editing~\cite{ILVR_Choi_2021,sdedit_meng_2022,edict_wallace_2022,egsde_zhao_2022}.
ILVR~\cite{ILVR_Choi_2021} refines a sample by adding the residual between the sample and the perturbed source image through a low-pass filter.
SDEdit~\cite{sdedit_meng_2022} starts the generation process from the noisy source image $y_M \sim q_{M|0}(y_M|x_0)$, where $M$ represents a middle time between $0$ and $T$, and is chosen to preserve the original overall structure and discard local details.
EGSDE~\cite{egsde_zhao_2022} employs an energy function pretrained on both the source and target domains to guide the inference process of a pretrained SDE.
EDICT~\cite{edict_wallace_2022} draws inspiration from affine coupling layers.

\begin{figure*}[t]
\centering
\includegraphics[width=1.0\linewidth]{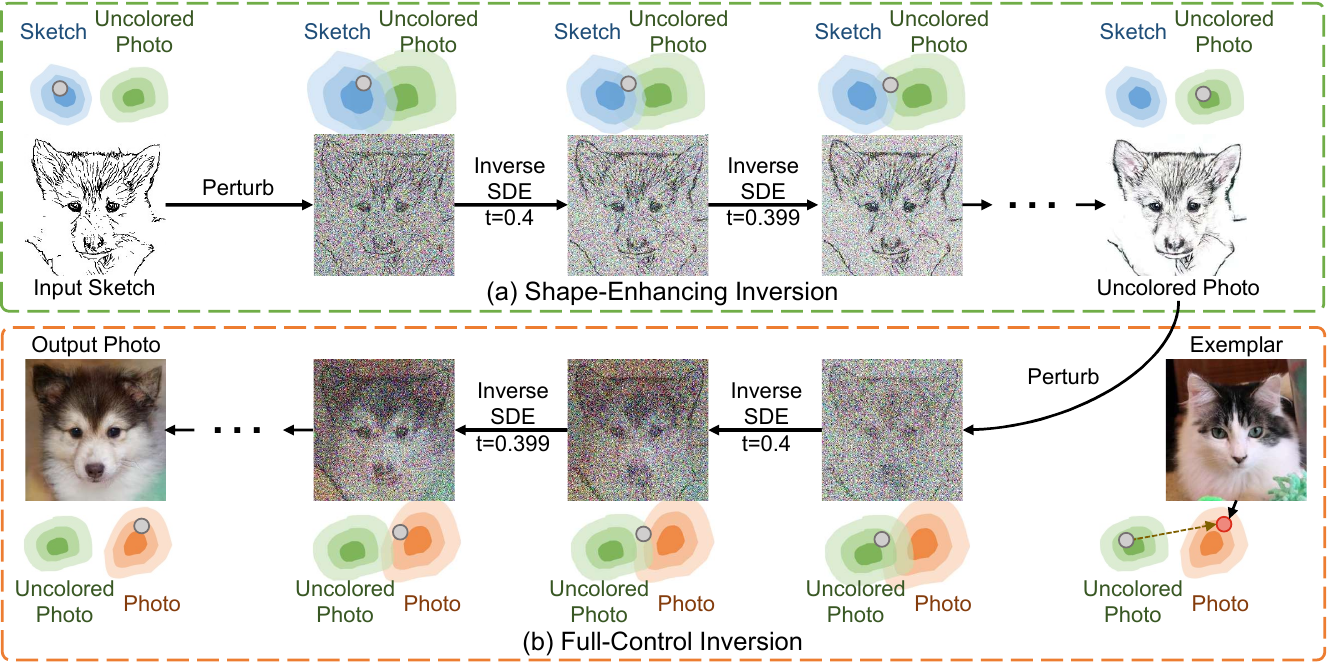}
\caption{
Overview of our Inversion-by-Inversion Translation via SDE. The blue, green, and orange contour plots represent the distributions of sketch, uncolored photo, and photo, respectively. The movement of the grey dot in the distribution denotes the sketch-to-photo synthesis process of our proposed Inversion-by-Inversion method. In our shape-enhancing inversion step (a), we first perturb the input sketch with the forward process of SDE. Then the inversion process of SDE will gradually remove the noise, and the uncolored photo is synthesized with the shape of the input sketch. During this procedure, we propose the shape-energy function to maintain the structure of the input sketch. After that, we perform the full-control inversion step (b) by first perturbing the uncolored photo and then using SDE inversion to denoise it. During this procedure, we use both the shape-energy function and the appearance-energy function for maintaining the structure of the input sketch and add the appearance (\textit{i.e.,} texture and color) from the given exemplar into the output photo (Best viewed in color).
}
\label{fig:overview}
\end{figure*}

\section{Method}

\subsection{Preliminaries: Stochastic Differential Equation}
The solution of Stochastic Differential Equations (SDE) is a time-varying random variable, denoted as $\boldsymbol{y}_t \in \mathbb{R}^D$, where $t \in [0, 1]$ represents time.
Following \cite{scorebased_song_2021}, we assume that $\boldsymbol{y}_0 \sim p_0 = p_{data}$ denotes a sample from the data distribution and a forward SDE produces $\boldsymbol{y}_t$ for $t \in \left(0,1\right]$ via a Gaussian diffusion.
\begin{equation}
\boldsymbol{y}_t = \alpha(t) \boldsymbol{y}_0 + \sigma(t) \boldsymbol{z}, \;
\boldsymbol{z} \sim \mathcal{N} \left(0, I\right),
\end{equation}
\noindent in which $\alpha(t)$ and $\sigma(t)$ are scalar functions that describe the magnitude of the data and noise, respectively. The probability density function of $y_t$ is denoted as $p_t$, $y_t \sim q_{t|0}(y_t|y_0)$.

The generative process of SDEs~\cite{scorebased_song_2021} begins with an initial noise vector $\left( \textit{i.e.}, \boldsymbol{y}_M \sim q_{M|0} \left( \boldsymbol{y}_M | \boldsymbol{x}_0 \right) \right)$ and performs iterative denoising (typically with a guidance signal, for example, in the form of text-conditional denoising), ending with a realistic image sample $x_0$. 
To solve the image editing problem~\cite{sdedit_meng_2022,egsde_zhao_2022}, it is necessary to run the reverse of this generative process. Formally, this problem is known as ``inversion", \textit{i.e.}, which involves finding the initial noise vector that produces the input image when passed through the diffusion process.

Let $x^{sk} \sim p^{sk}_0 = p_{sketch}$ denote the sketch data distribution on $\mathbb{R}^{D}$ and $x^{ex} \sim p^{ex}_0 = p_{exemplar}$ denote the exemplar data distribution on $\mathbb{R}^{D}$. Given the input sketch $x^{sk}$ in the sketch domain $\mathcal{A}$ and the exemplar image $x^{ex}$ in the exemplar domain $\mathcal{B}$, the goal of our method is to generate photos with the appearance of the given exemplar and the shape of the input sketch by utilizing a pre-trained SDE. 
Similar to prior work~\cite{sdedit_meng_2022,egsde_zhao_2022}, we employ the SDE trained solely in the target domain as defined in Eq.~\ref{eq:reverse_sde}, which defines a marginal distribution of the target images and primarily contributes to the realism of the transferred samples.

\subsection{Inversion by Inversion Translation via SDE}
As shown in Figure~\ref{fig:overview}, we propose an \textit{Inversion-by-Inversion} Sketch-to-Photo synthesis method via SDE, a single framework for disentangling conditional input and exemplar as well as synthesizing high-fidelity images, which consists of two inversion steps: 
(a) Shape-Enhancing Inversion: In this inversion step, we enhance the shape of the input sketch and generate an uncolored photo by using the shape-energy function to maintain the geometric structure of the sketch during the inverse SDE process.
(b) Full-Control Inversion: Using the uncolored photo as input, we extract the visual details (\textit{i.e.,} color and texture) from the exemplar and add them to uncolored photos using the shape-energy function and appearance-energy function during the inverse-SDE process, resulting in the final output.
Details of our method are described as follows.

\subsection{Shape-Enhancing Inversion}
As it is challenging to preserve the geometric structure of the sketch when directly adding the appearance of the exemplar, we propose shape-enhancing inversion for generating uncolored photos via SDE in the first stage.
The procedure of the shape-enhancing inversion is illustrated in Figure~\ref{fig:overview}(a). The input sketch is first perturbed by uniform noise and then the inverse SDE is used for denoising. During this inversion step, the grey dot (\textit{i.e.,} current state of the image) gradually moves from the sketch domain to the uncolored photo domain. By using our proposed shape-energy function, our shape-enhancing inversion based on SDE generates uncolored photos that have stronger shape information and can better preserve the shape during the next full-control inversion step.

We first define a valid conditional distribution $p\left(y^{\prime}_0|x^{sk}_0\right)$ by compositing a pre-trained SDE and a pre-trained energy function under mild regularity conditions as follows:
\begin{equation}
\text{d}\boldsymbol{y^{\prime}}=\left[
\mathbf{f}(\boldsymbol{y}^{\prime}_0,t) 
- g(t)^2 \mathbf{s}(\boldsymbol{y}^{\prime}_0,t) 
- \lambda_1 \mathcal{E}_{g} 
\right]\text{d}t 
+ g(t) \text{d}\overline{\boldsymbol{\omega}}
\label{eq:reverse_dse_energy_in1}
\end{equation}
\noindent where $\overline{\boldsymbol{\omega}}$ is a reverse-time standard Wiener process, $\text{d}t$ is an infinitesimal negative timestep, $\boldsymbol{s} \left(\cdot, \cdot\right) : \mathbb{R}^{D} \times \mathbb{R} \xrightarrow{} \mathbb{R}^D$ is the score-based model in the pre-trained SDE and $\mathcal{E}_g : \mathbb{R}^D \times \mathbb{R}^D \times \mathbb{R} \xrightarrow{} \mathbb{R}$ is the shape-energy function.
The start point $\boldsymbol{y}_M$ is sampled from the perturbation distribution $q_{M|0} \left( \boldsymbol{y}_M | \boldsymbol{x}_0 \right)$, where $M=0.4T$ typically.
Then we obtain the uncolored photo by taking the samples at endpoint $t = 0$ following the SDE in Eq.~\ref{eq:reverse_dse_energy_in1}. 

To perform shape-enhancing inversion based on energy-guided SDE, we design a new shape-energy function, which is formulated as follows:
\begin{equation}
\hspace{-1mm} 
\boldsymbol{\mathcal{E}}_g
=
\boldsymbol{\mathcal{E}} \left( \boldsymbol{y}^{\prime}, \boldsymbol{x}^{sk}_0, t \right) 
= 
\boldsymbol{\lambda}_g \boldsymbol{\mathbb{E}}_{q_{t|0}(x^{sk}_t|x^{sk})} \mathcal{S}_g \left( \boldsymbol{y}^{\prime}, \boldsymbol{x}^{sk}_t, t \right) 
\label{eq:shape_energy}
\end{equation}
\noindent where $\mathcal{E}_g \left(\cdot,\cdot,\cdot \right): \mathbb{R}^{D} \times \mathbb{R}^{D} \times \mathbb{R} \xrightarrow{} \mathbb{R}$, $\boldsymbol{x}^{sk}_t$ is the perturbed conditional images (\textit{i.e.}, sketch) in the forward SDE, $q_{t|0} (\cdot | \cdot)$ is the perturbation kernel from time $0$ to time $t$ in the forward SDE, $\mathcal{S}_g \left(\cdot, \cdot, \cdot \right): \mathbb{R}^D \times \mathbb{R}^D \times \mathbb{R} \xrightarrow{} \mathbb{R}$ is function measuring the similarity between the sketch of sample and perturbed sketch input, and $\lambda_g \in \mathbb{R}_{>0}$ is the weighting hyper-parameter.

The objective of $\mathcal{S}_g \left(\cdot, \cdot, \cdot\right)$ is to compute the distance between the sketches derived from generated photos and the input sketches, which encourages the shape of generated photos to be aligned with the input sketches. 
To achieve this, we introduce a pre-trained photo-to-sketch network~\cite{infodrawing_Chan_2022} $\Phi_g \left(\cdot\right): \mathbb{R}^{C \times H \times W} \xrightarrow{} \mathbb{R}^{C \times H \times W}$, which translates a photo into a sketch, where $C$ is the channel-wise dimension, $H$ and $W$ are height and width.
Building upon the pre-trained photo-to-sketch network, $\mathcal{S}_g \left(\cdot, \cdot, \cdot\right)$ is formulated as follows:
\begin{equation}
\mathcal{S}_g \left( \boldsymbol{y}^{\prime}, \boldsymbol{x}^{sk}_t, t \right) = \Vert \Phi_g \left(\boldsymbol{y}^{\prime}\right) - \boldsymbol{x}^{sk}_t \Vert_2^2
\end{equation}

\begin{algorithm}[t]
\caption{\textit{Inversion-by-Inversion} Translation via SDE.}\label{algo:ours_method}
\begin{algorithmic}[1]
\Require {\bfseries Input:} the sketch input $x^{sk}_0$, the initial time $M$, denoising steps $N$, weighting hyper-parameters $\lambda_s$, $\lambda_a$, the similarity function $\mathcal{S}_g (\cdot,\cdot)$, $\mathcal{S}_a (\cdot,\cdot)$, the score function $s(\cdot,\cdot)$, total number of inversion stages $I=2$.
\State\textbf{Initialize:} $\boldsymbol{y} \sim q_{M|0} \left( \boldsymbol{y}_M | \boldsymbol{x}^{sk}_0 \right)$. \Comment{the start point}
\State $h = \frac{M}{N}$ 
\For{$i=N$ {\bfseries to} $1$}  \Comment{shape-enhancing inversion}
\State $s \xrightarrow{} i h$.
\State $\boldsymbol{\mathcal{E}}_g \xrightarrow{} \boldsymbol{\mathcal{E}} \left( \boldsymbol{y}, \boldsymbol{x}^{sk}_0, s \right)$
\State $\text{d}\boldsymbol{y}=\left[
\boldsymbol{f}(\boldsymbol{y},s) 
- g(t)^2 \boldsymbol{s}(\boldsymbol{y},s) - \mathcal{E}_{g} \right]\text{d}t + g(t) \text{d}\overline{\boldsymbol{\omega}}$
\State $\boldsymbol{z} \sim \mathcal{N}(0, I) \text{\:if\:} i > 1, \text{else\:} \boldsymbol{z}=0$
\State $\boldsymbol{y} \xleftarrow{} \boldsymbol{y} + g(s) \sqrt{h} \boldsymbol{z}$
\EndFor
\State $\boldsymbol{y}^{\prime} \sim q_{M|0} \left( \boldsymbol{y}^{\prime}_M | \boldsymbol{y} \right)$ \Comment{sample perturbed output of shape-enhancing inversion}
\For{$i=N$ {\bfseries to} $1$}  \Comment{full-control inversion}
\State $s \xrightarrow{} i h$.
\State $\boldsymbol{x^{ex}} \sim q_{s|0}(\boldsymbol{x^{ex}}_M | \boldsymbol{x^{ex}_0})$ \Comment{sample perturbed exemplar from the perturbation kernel}
\State $\boldsymbol{\widetilde{\mathcal{E}}} \xrightarrow{} \lambda_g \boldsymbol{\mathcal{E}_g} \left( \boldsymbol{y}^{\prime}, \boldsymbol{x}^{sk}_0, s \right) - \lambda_a \boldsymbol{\mathcal{E}_a}(\boldsymbol{y}^{\prime}, \boldsymbol{x}^{ex}, s)$ 
\State $\boldsymbol{z} \sim \mathcal{N}(0, I) \text{\:if\:} i > 1, \text{else\:} \boldsymbol{z}=0$
\State $\boldsymbol{y}^{\prime} \xleftarrow{} \boldsymbol{y}^{\prime} + g(s) \sqrt{h} \boldsymbol{z}$
\EndFor
\State $\boldsymbol{y}_0 \xleftarrow{} \boldsymbol{y}$
\State \textbf{return} $\boldsymbol{y}_0$ 
\end{algorithmic}
\end{algorithm}

\subsection{Full-Control Inversion}

After generating the uncolored photos from the shape-enhancing inversion stage, we aim to add the appearance (\textit{i.e.,} color and texture) from the given exemplar to the uncolored photo while maintaining the geometric structure of the input sketch. As shown in the example of Figure~\ref{fig:overview}, if we take a cat photo as an exemplar, our model aims to transfer the color and texture of the cat's fur into the final generated dog photo accordingly. 

The procedure of the full-control inversion stage is illustrated in Figure~\ref{fig:overview}(b). In this stage, we first perturb the generated uncolored photo with Gaussian noise. We then perform the inverse SDE procedure for denoising. During the denoising procedure, the gray dot (\textit{i.e.,} the current state of the image) gradually moves from the uncolored photo domain to our target photo domain, guided by  our proposed energy function $\boldsymbol{\mathcal{E}} \left( \boldsymbol{y}^{\prime}, \boldsymbol{x}^{sk}_0, \boldsymbol{x}^{ex}_0, t \right)$, and the final output photo is generated. 
In this inversion step, to transfer the texture and color from the exemplar, we redefine the valid conditional distribution $p\left(y^{\ast}_0|y^{\prime}_0,x^{ex}_0\right)$ by introducing exemplar as follows:
\begin{equation}
\begin{split}
&\text{d}\boldsymbol{y^{\ast}} = \left[
\mathbf{f}(\boldsymbol{y}^{\prime},t) 
- g(t)^2 \mathbf{s}(\boldsymbol{y}^{\prime},t) 
- \widetilde{\mathcal{E}}
\right]\text{d}t 
+ g(t) \text{d}\overline{\boldsymbol{\omega}}
\\& \text{ s.t. } 
\widetilde{\mathcal{E}} = \lambda_2 \left( 
\nabla_{y^{\ast}} \mathcal{E}_g(\boldsymbol{y}^{\prime}, x^{sk}_0, t) 
+ \nabla_{y^{\ast}} \mathcal{E}_a(\boldsymbol{y}^{\prime}, x^{ex}_0, t)
\right)
\label{eq:reverse_dse_energy_in2}
\end{split}
\end{equation}
\noindent where $\widetilde{\mathcal{E}}$ is our newly proposed energy function such that it encourages the sample to preserve the sketch's geometric structure and transfers the content of the exemplar.

We decompose the energy function $\boldsymbol{\mathcal{E}} \left( \boldsymbol{y}^{\prime}, \boldsymbol{x}^{sk}_0, \boldsymbol{x}^{ex}_0, t \right)$ as the sum of two log potential functions~\cite{lEBM_gao_2021}:
\begin{equation}
\begin{split}
\hspace{-1mm} 
\boldsymbol{\mathcal{E}} \left( \boldsymbol{y}^{\prime}, \boldsymbol{x}^{sk}_0, \boldsymbol{x}^{ex}_0, t \right) 
&= 
\boldsymbol{\mathcal{E}}_g \left( \boldsymbol{y}^{\prime}, \boldsymbol{x}^{sk}_0, t \right) 
+
\boldsymbol{\mathcal{E}}_a \left( \boldsymbol{y}^{\prime}, \boldsymbol{x}^{ex}_0, t \right) 
\\& =
\boldsymbol{\lambda}_g \boldsymbol{\mathbb{E}}_{q_{t|0}(x^{sk}_t|x^{sk})} \mathcal{S}_g \left( \boldsymbol{y}^{\prime}, \boldsymbol{x}^{sk}_t, t \right) 
\\& + 
\boldsymbol{\lambda}_a \boldsymbol{\mathbb{E}}_{q_{t|0}(x^{ex}_t|x^{ex})} \mathcal{S}_a \left( \boldsymbol{y}^{\prime}, \boldsymbol{x}^{ex}_t, t \right),
\end{split}
\end{equation}
\noindent where both $\mathcal{E}_g \left(\cdot,\cdot,\cdot \right) : \mathbb{R}^{D} \times \mathbb{R}^{D} \times \mathbb{R} \xrightarrow{} \mathbb{R}$ and $\mathcal{E}_a \left(\cdot,\cdot,\cdot \right) : \mathbb{R}^{D} \times \mathbb{R}^{D} \times \mathbb{R} \xrightarrow{} \mathbb{R}$ are the log potential functions, $\boldsymbol{x}^{sk}_t$ is the perturbed conditional images (\textit{i.e.}, sketch) in the forward SDE, $q_{t|0} (\cdot | \cdot)$ is the perturbation kernel from time $0$ to time $t$ in the forward SDE, $\mathcal{S}_g \left(\cdot, \cdot, \cdot \right) : \mathbb{R}^D \times \mathbb{R}^D \times \mathbb{R} \xrightarrow{} \mathbb{R}$ and $\mathcal{S}_a \left(\cdot, \cdot, \cdot \right) : \mathbb{R}^D \times \mathbb{R}^D \times \mathbb{R} \xrightarrow{} \mathbb{R}$ are two functions measuring the similarity between the sample and perturbed source image, and $\lambda_g \in \mathbb{R}_{>0}$, $\lambda_a \in \mathbb{R}_{>0}$ are two weighting hyper-parameters.

During the full-control inversion process, we use the same shape-energy function $\mathcal{E}_g \left(\cdot, \cdot, \cdot\right)$ as used in shape-enhancing inversion stage to maintain the shape of the input sketch. Moreover, we  introduce a hierarchical feature extractor to guide appearance reconstruction.
Specifically, we divide the appearance control into two levels: the pixel level and the feature level. 
At the pixel level, we apply a non-trainable low-pass filter~\cite{ILVR_Choi_2021} with a large window size to add color information to the generated samples. 
At the feature level, reconstructing appearance from the exemplar is not as straightforward as aligning colors in RGB pixels, as photos may exhibit distinct styles. Therefore, we propose to use the CLIP Visual Encoder~\cite{CLIP_Radford_2021,vinker2022clipasso} features to enable appearance control during the full-control inversion stage. The similarity function $\mathcal{S}_a \left(\cdot,\cdot,\cdot\right)$ is formulated as
\begin{equation}
\begin{split}
\mathcal{S}_a \left( \boldsymbol{y}^{\prime}, \boldsymbol{x}^{sk}_t, t \right) 
& = \Vert \Omega_N \left(\boldsymbol{y}^{\prime}\right) - \Omega_N \left(\boldsymbol{x}^{ex}_t\right) \Vert_2^2
\\& +
\sum_{i=0}^L \Vert \Psi_i \left(\boldsymbol{y}^{\prime}\right) - \Psi_i \left(\boldsymbol{x}^{ex}_t\right) \Vert_2^2
\end{split}
\label{eq:energy_texture_metric}
\end{equation}
in which $\Psi_i$ represent the CLIP Visual Encoder~\cite{CLIP_Radford_2021} network of the 1th and 4th layers~\cite{vinker2022clipasso} and $\Omega_N (\cdot)$ denote a linear low-pass filtering operation with a factor of $N$. In this work, $N$ is set to be 64. Note that we also use negative squared $L_2$ distance as the similarity metric in both pixel-level and feature-level. 

\begin{table*}[t]
\centering
\caption{
Quantitative comparison. ILVR~\cite{ILVR_Choi_2021}, SDEdit~\cite{sdedit_meng_2022} and EGSDE~\cite{egsde_zhao_2022} are reproduced using the released code. Our proposed method uses the default-parameters ($\lambda_g = 0.1$, $\lambda_a = 30$, $M = 0.4T$). The $L_2$ distance represents the structural similarity between the sketch of the generated photos and the input sketches. PSNR evaluates the similarity between the generated photos and the exemplars.
}
\resizebox{44em}{!}{
\begin{tabular}{l|cc|ccc|ccc}
\toprule
\multirow{2}{*}{Method} & \multicolumn{2}{c|}{Effect Control} & \multicolumn{3}{c|}{Cat $\xrightarrow{}$ Dog} & \multicolumn{3}{c}{Wild$\xrightarrow{}$Dog} \\

 & Shape & Texture & FID $\downarrow$ & $L_2$ $\downarrow$ & PSNR $\uparrow$ & FID $\downarrow$ & $L_2$ $\downarrow$ & PSNR $\uparrow$  \\
\midrule
AODA~\cite{xiang2022adversarial} & $\checkmark$ & $\times$ & 183.75 & 70.52 & 9.55 & 217.87 & 52.87 & 10.40  \\ 

ILVR~\cite{ILVR_Choi_2021} & $\times$ & $\checkmark$ & 157.36 & 38.06 & 5.16 & 157.36 & 38.06 & 5.20  \\ 
ILVR (Mixup1)~\cite{ILVR_Choi_2021} & $\times$ & $\checkmark$ & 157.00 & 39.02 & 6.37 & 157.60 & 38.51 & 6.40 \\ 
ILVR (Mixup2)~\cite{ILVR_Choi_2021} & $\times$ & $\checkmark$ & 131.04 & 43.68 & 8.34 & 138.34 & 41.65 & 8.31 \\ 

SDEdit~\cite{sdedit_meng_2022} & $\times$ & $\checkmark$ & 103.62 & \textbf{36.74} & 5.39 & 103.62 & \textbf{36.74} & 5.43  \\
SDEdit (Mixup1)~\cite{sdedit_meng_2022} & $\times$ & $\checkmark$ & 107.98 & 43.87 & 8.26 & 111.15 & 41.91 & 8.23 \\
SDEdit (Mixup2)~\cite{sdedit_meng_2022} & $\times$ & $\checkmark$ & 83.42 & 58.67 & 14.47 & 79.58 & 54.84 & 14.11\\

EGSDE~\cite{egsde_zhao_2022} & $\times$ & $\checkmark$ & 126.03 & 39.29 & 5.41 & 126.03 & 39.29 & 5.44 \\ 
EGSDE  (Mixup1)~\cite{egsde_zhao_2022} & $\times$ & $\checkmark$ & 114.80 & 44.31 & 8.23 & 114.43 & 42.39 & 8.21\\ 
EGSDE  (Mixup2)~\cite{egsde_zhao_2022} & $\times$ & $\checkmark$ & 72.57 & 57.28 & 14.15 & 68.13 & 52.36 & 13.75 \\ 

DiffSketching~\cite{Wang_2022_DiffSketching} & $\checkmark$ & $\times$ & 110.50 & 41.21 & 8.84 & 121.47 & 41.72 & 9.57 \\ 
DiSS~\cite{cheng_2023_adaptively} & $\checkmark$ & $\checkmark$ & 122.42 & 64.62 & \textbf{15.94} & 135.11 & 63.55 & \textbf{15.58} \\ 
\midrule
Ours & $\checkmark$ & $\checkmark$ & \textbf{31.26}  & 43.70 & 13.07 & \textbf{27.11} & 43.11 & 14.99 \\ 
\bottomrule
\end{tabular}
}
\label{tab:quality_results}
\end{table*}

\section{Experiments}

\begin{figure*}[t]
\centering
\includegraphics[width=0.72\linewidth]{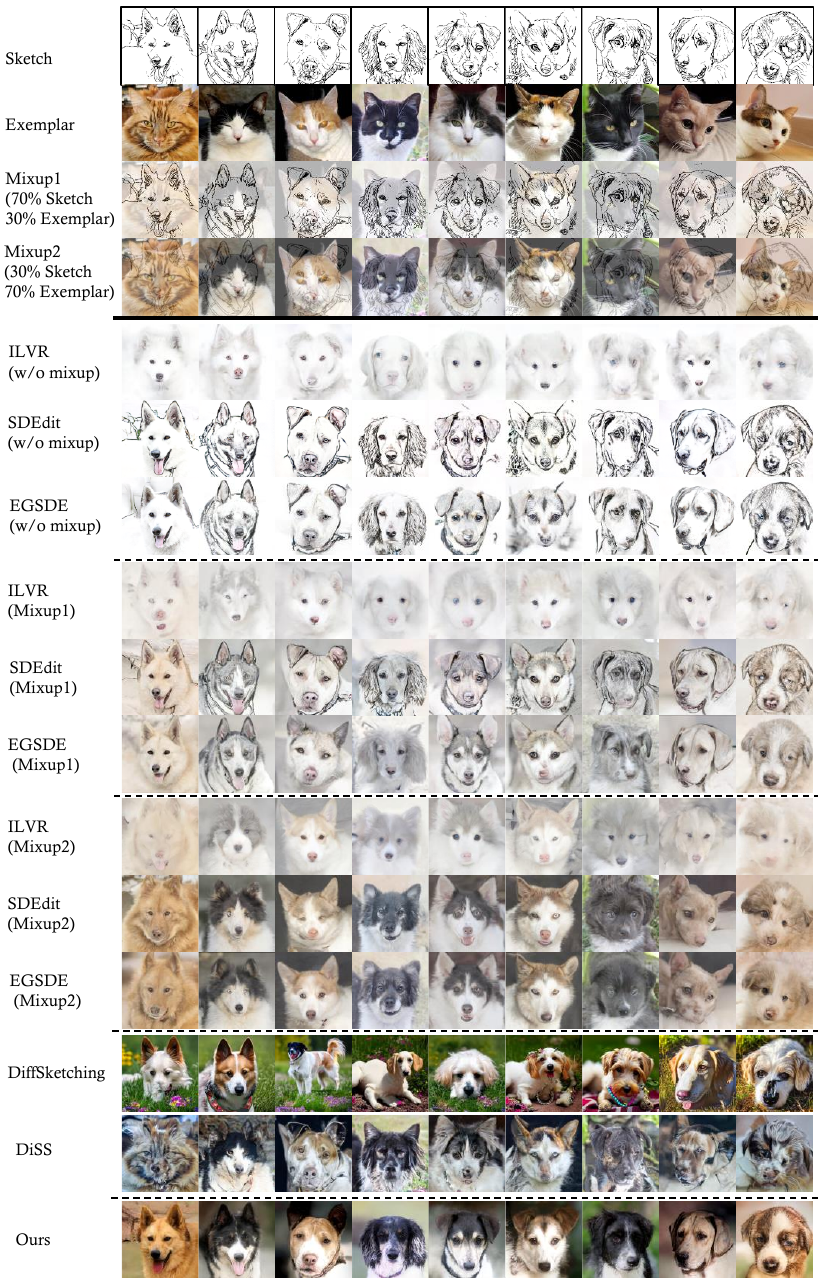}
\caption{
Comparison of the results of different methods on Cat $\xrightarrow{}$ Dog. Considering ILVR, SDEdit and EGSDE cannot simultaneously take two images as input, we conducted two versions of experiments for these methods: using the sketch directly as input or mix up the sketch and exemplar as a single image and use it as the input.
“Mixup 1” denotes a blended image of
70\% sketch and 30\% exemplar, and “Mixup 2” represents a
blended image of 30\% sketch and 70\% exemplar. It is challenging for these methods to achieve a balance between the shape control and appearance control. 
Among all methods, our method achieves the highest visual quality and faithfulness.
}
\label{fig:regular_sketch_compare}
\end{figure*}

\subsection{Experimental Setup}

\subsubsection{Datasets}
We evaluate our proposed method on the AFHQ~\cite{StarGAN_Choi_2020} dataset. All images are resized to $256 \times 256$. The AFHQ dataset consists of three domains: ``cat" with 5153 training images, ``dog" with 4739 training images of ``dog", and ``wild" (\textit{e.g.} tiger, lion, wolf, etc) with 4738 training images. Each domain also has 500 testing images. We evaluate the Cat $\xrightarrow{}$ Dog translation and the Wild $\xrightarrow{}$ Dog translation on this dataset, which means the exemplar images are from ``cat" or ``wild" domain, while the input sketches are from the ``dog" domain.

\subsubsection{Implementation Details}
To achieve $\mathcal{S}_g(\cdot,\cdot,\cdot)$ in the shape-enhancing inversion, we use a pretrained photo-to-sketch network~\cite{infodrawing_Chan_2022} $\Phi_g \left(\cdot\right)$ to predict edge maps. Specificity, the pre-trained photo-to-sketch network~\cite{infodrawing_Chan_2022} consists of an encoder-decoder architecture with ResNet blocks in the middle and a patch-based discriminator~\cite{condi_gan_isola_2017}.
Besides, we introduce a linear low-pass filtering operation $\Omega_N (\cdot)$~\cite{ILVR_Choi_2021}, a sequence of downsampling and upsampling operations with a factor of $N = 64$. Unlike ILVR, we get low frequency information about the color area by setting a large value of $N$ and discarding high frequency information such as shapes and textures. 
To control texture representation in Eq.~\ref{eq:energy_texture_metric}, we introduce the CLIP Visual Encoder~\cite{CLIP_Radford_2021} $\Psi_i$ by obtaining the output of the 0th and 3rd layers~\cite{vinker2022clipasso}.

For the generation process, the weight parameters $\lambda_g$ and $\lambda_a$ are set to be $0.1$ and $50$, respectively, by default. The initial time $M$ and denoising steps are initially set as $0.4T$ and $400$  respectively, by default.
During the sampling procedure, all photos are resized to $256 \times 256$ and the values are normalized to $\left[ -1,1 \right]$.
For Cat $\xrightarrow{}$ Dog and Wild $\xrightarrow[]{}$ Dog translations on the AFHQ dataset, we use the pre-trained score-based diffusion model (SBDM) provided in the official code of ILVR~\cite{ILVR_Choi_2021}. The pre-trained model includes the variance and mean networks but we only use the mean network.

Below, we provide additional hyperparameters and implementation details for each experiment. We use the checkpoints of publicly available pretrained SDE provided by~\cite{ILVR_Choi_2021,egsde_zhao_2022}.

\noindent\textbf{Cat to Dog.}\quad In the shape-enhancing inversion, we set $K = 1, M = 0.4T, N = 400$, and the weight parameter $\lambda_g$ to  $0.1$.
In the full-control inversion, we set $K = 1, M = 0.4T, N = 400$, the weight parameters $\lambda_g$ and $\lambda_a$ to $0.1$ and $2$, respectively.
We observed that the values of $K$ between $1$ and $3$ work reasonably well, with larger values of $K$  generating more realistic images but at a higher computational cost.

\noindent\textbf{Wild to Dog.}\quad In this experiment, we use the same hyperparameters as those in the Cat2Dog experiment.

\begin{figure*}[t]
\centering
\includegraphics[width=1\linewidth]{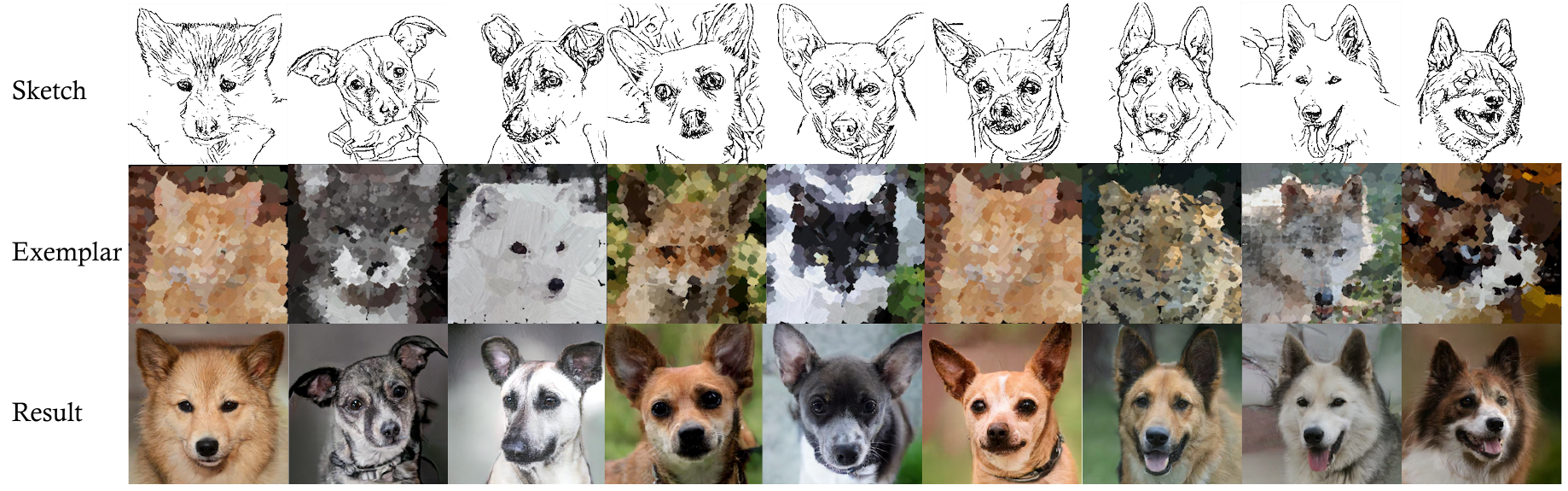}
\caption{
Stroke-based sketch-to-photo synthesis with our Inversion-by-Inversion method on Cat $\xrightarrow{}$ Dog.
}
\label{fig:stroke_results}
\end{figure*}

\noindent\textbf{Stroke-based Image Synthesis.}\quad 
We use the strokes produced by Painter Transformer~\cite{liu2021paint} as the exemplar, as shown in the fifth column of Figure~\ref{fig:all_in_one} and Figure~\ref{fig:stroke_results}.
In the shape-enhancing inversion, we set $K = 1, M = 0.4T, N = 400$, and the weight parameter $\lambda_g$ to $0.2$.
For the full-control inversion, we set $K = 1, M = 0.4T, N = 400$, the weight parameters $\lambda_g$ and $\lambda_a$ to $0.2$ and $1$, respectively.

\noindent\textbf{Segmentation-based Image Synthesis.}\quad 
As shown in the sixth column of Figure~\ref{fig:all_in_one}, we use the segmentation map as the exemplar, and we use the same hyperparameters as those in the Stroke-based Image Synthesis experiment.

\begin{figure*}[h]
\centering
\includegraphics[width=0.7\linewidth]{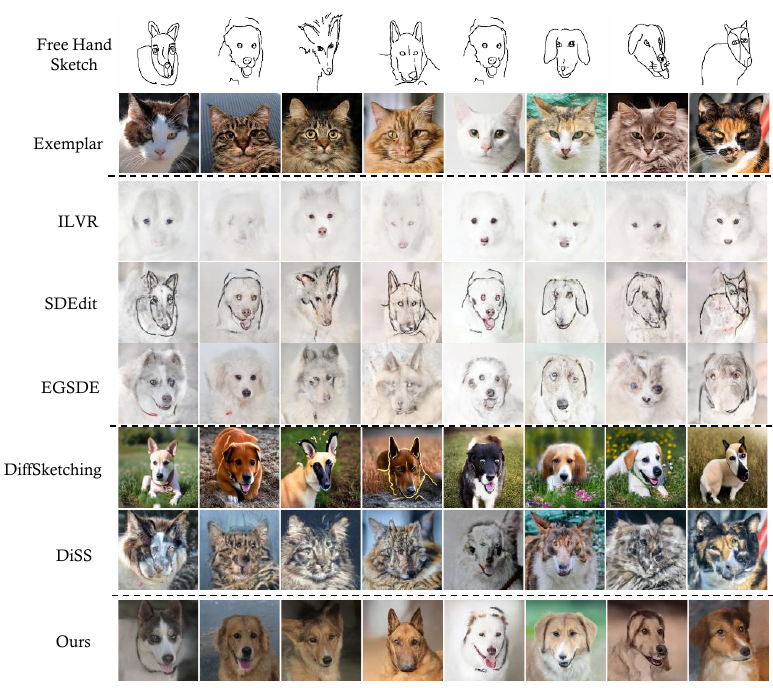}
\caption{
Comparison of the results when using freehand sketches for shape control. The sketches are sampled from the sketchy~\cite{sketchy} dataset. Considering ILVR, SDEdit and EGSDE cannot simultaneously take the sketch and exemplar as input, we compare to use the combined images of sketches and exemplars (``Mixup1") as the input of ILVR, SDEdit and EGSDE.
}
\label{fig:free_hands_sketch}
\end{figure*}

\noindent\textbf{Freehand sketch-based Image Synthesis.}\quad 
We use the data samples from the Sketchy dataset as exemplar, as shown in the last three columns of Figure ~\ref{fig:all_in_one} and Figure ~\ref{fig:free_hands_sketch}.
In the shape-enhancing inversion, we set $K = 1, M = 0.5T, N = 500$, and the weight parameters $\lambda_g$ to $0.01$.
In the full-control inversion, we set $K = 1, M = 0.5T, N = 500$, the weight parameters $\lambda_g$ and $\lambda_a$ to $0.01$ and $2$, respectively.

\begin{figure}[h]
\centering
\includegraphics[width=1.0\linewidth]{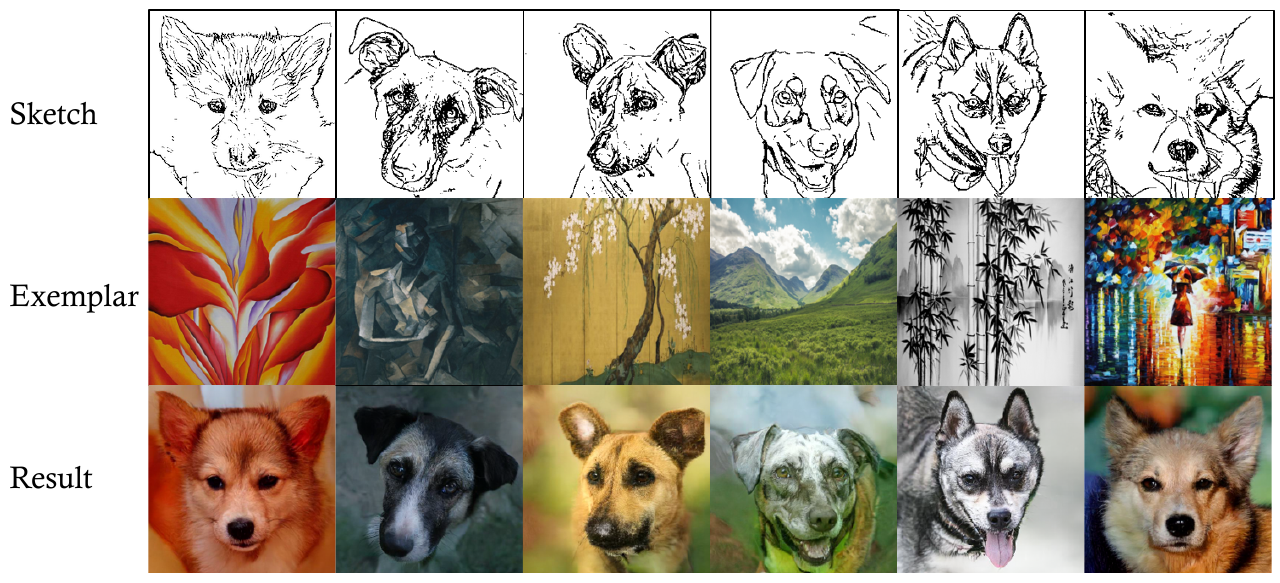}
\caption{
Style-based sketch-to-photo synthesis with our Inversion-by-Inversion method on Cat $\xrightarrow{}$ Dog. 
The figure in this context consists of artistic images, and our method effectively translates them.
}
\label{fig:style_results}
\end{figure}

\subsubsection{Evaluation Metrics}
We evaluate the generation results based on the criteria of \textit{realism} and \textit{fidelity of texture and shape}. (1) To quantify realism, we use the Fréchet Inception Distance (FID)  metric using the Pytorch-FID code on the AFHQ dataset. The FID score measures the visual quality and distance between the distributions of real and generated images. Following CUT~\cite{cut_park_2020}, we use the test data as reference without any data preprocessing.
(2) To quantify shape fidelity, we report the $L_2$ distance summed over all pixels between the  input sketch and the sketch derived from the sampled images. Given the sparse nature of the sketch, direct calculation of pixel-level $L_2$ distance results in relatively large values. To address this issue, we apply pixel-level normalization to the calculation results. 
(3) Style relevance is evaluated based on the Peak Signal-to-Noise Ratio (PSNR) metric between the input exemplar and the sketch of the generated images.

\subsubsection{Training and Inference Time}
In our proposed \textit{Inversion-by-Inversion} method, all the networks including the diffusion model, the photo-to-sketch network $\Phi_g \left(\cdot\right)$ and the CLIP Visual Encoder $\Psi_i \left(\cdot\right)$ are pre-trained, and we directly use their published checkpoints. Therefore, we do not require extra training steps for these sub-modules.
Regarding the inference time for the Cat $\xrightarrow{}$ Dog experiment, sampling a photo takes approximately 30 seconds on an NVIDIA 3090 GPU.

\subsection{Quantitative Results}

\noindent\textbf{Baseline Methods.} We compare our proposed method with three recently proposed state-of-the-art SDE-based image translation methods: ILVR~\cite{ILVR_Choi_2021}, SDEdit~\cite{sdedit_meng_2022}, EGSDE~\cite{egsde_zhao_2022}, DiSS~\cite{cheng_2023_adaptively}, and DiffSketching~\cite{Wang_2022_DiffSketching}, which serve as  diffusion-model baseline method for comparison. We also compare a GAN-based method AODA~\cite{xiang2022adversarial}.

The quantitative results are provided in Table~\ref{tab:quality_results}.  Our approach achieves the best FID score and outperforms other baseline methods by a large margin. For example, in the Cat$\xrightarrow{}$Dog translation, our method achieves an FID score of $31.26$, while the second best FID is  $53.63$ by EGSDE. These results demonstrate that our method generates more photo-realistic images, and the distribution of our generated images is closer to the real distribution when compared with other baseline methods. In terms of $L_2$ distance between the input sketch and the sketch of the output photo, our method  outperforms SDEdit, EGSDE, and Diss, but slightly worse than ILVR and DiffSketching. It is worth noting that the FID of these two methods are much worse than our method. For example, our method achieves an FID score of $27.11$ on the Wild$\xrightarrow{}$Dog translation, while the corresponding result of ILVR is $113.94$. The PSNR results evaluate the similarity between the generated results and the exemplar. Our method achieves better results than most methods, and comparable results with DiSS. These results demonstrate that our proposed method is capable of producing images with the highest visual quality while preserving the geometric structure of the input sketch and the appearance of the exemplar.

\subsection{Qualitative Results}

Next, we compare the baseline methods with our method in terms of their qualitative results.

\subsubsection{Comparison with Baseline Methods}

In Figure~\ref{fig:regular_sketch_compare}, we illustrate the generation results of different methods based on regular sketches and different types of exemplars. 
Three baseline methods, including ILVR, SDEdit, EGSDE can only accept a single input image. Therefore, besides directly feeding sketches into these methods for translation, we attempted to combine a sketch and an exemplar by mixing them with a preset ratio and used it as the input for these methods.
In Figure~\ref{fig:regular_sketch_compare}, we first compare the results of different methods when directly taking a single sketch as input. It is obvious that the results of these baseline methods appear whitish as they did not obtain color or texture information.
When the input changed to the mixup version, i.e., ``Mixup1" and ``Mixup2", the visual quality of the results is improved,  but it is still challenging for these baseline methods to achieve a balance between shape control (shape) and exemplar control (appearance). ``Mixup 1" denotes a blended image of 70\% sketch and 30\% exemplar, and ``Mixup 2" represents a blended image of 30\% sketch and 70\% exemplar. 
When the input is ``Mixup1", the results of the baseline methods can preserve the shape of the original sketch well, but the colors appear blenched. When increasing the ratio of the exemplar images in the combined images, these baseline methods can produce results that are more aligned with the exemplar images in colors and textures, but the shape of the original sketch cannot be preserved. 
DiffSketching can produce realistic images, but it cannot faithfully reflect the shape of the original input sketches in  generated photos. Although DiSS achieves the highest PSNR, the visual quality of the generated photos are much worse than other methods. 

By contrast, our method can produce images that satisfy both control conditions simultaneously.

In Fig.~\ref{fig:free_hands_sketch}, we also compare the results of different methods when using freehand sketches for shape control. This is a more challenging task as freehand sketches exhibit more deformations and abstractions than regular sketches. The sketches used in our experiment are randomly sampled from the Sketchy Dataset~\cite{sketchy} and were drawn by amateurs.
As shown in Figure~\ref{fig:free_hands_sketch}, when simply taking a sketch as input, the baseline methods can only generate uncolored images. Furthermore, when taking a blended image of the sketch and the exemplar as the input, these methods are still struggling to generate a high-quality image. We can see that the generation results of ILVR (denoted as ``ILVR(mixup)") cannot maintain the structure suggested by the sketch, and the results of SDEdit and EGSDE cannot generate realistic images. The results of our method are much better than baseline methods in terms of faithfulness and realism, demonstrating its effectiveness.

\subsubsection{Using Style Images as Exemplars}
\label{exp:style_results}
In our method, we adopt the exemplar to control the appearance (\textit{i.e.,} color and texture) of generated images. These images can be real photos or style images (as shown in Figure~\ref{fig:all_in_one}). We provide more qualitative results of using style images as exemplar in Figure~\ref{fig:style_results}. As shown in Figure~\ref{fig:style_results}, the generated results accurately capture the colors of the style images.

\subsubsection{Using Stroke Images as Exemplars}
We also provide the generation results when simply using strokes as the exemplar in Fig.~\ref{fig:stroke_results}. We observe that the colors of the strokes can be faithfully added to our results. For example, in the last column, the white and the brown colors are accurately reflected in the results as desired. 

\subsection{Ablation Study}

\begin{figure}[h]
\centering
\includegraphics[width=\linewidth]{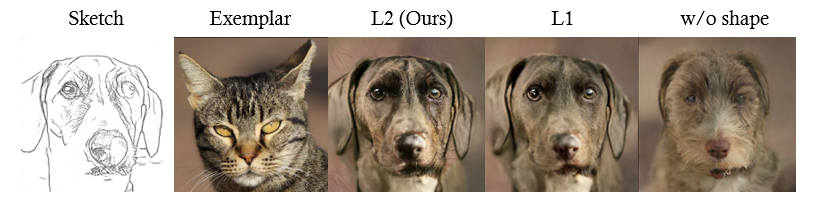}
\caption{
The effectiveness of the proposed shape-energy function. The shape-energy function is used to control the shape of the synthesized photo. The model without it (denoted as \textbf{w/o shape}) cannot synthesize a photo with the desired shape. Both $L_{2}$ and $L_{1}$ distance can preserve the shape of the input sketch, and we use $L_{2}$ in our final model as it empirically performs better.\vspace{-2mm}
}
\label{fig:ablation_geom_metric}
\end{figure}

\begin{figure}[h]
\centering
\includegraphics[width=1\linewidth]{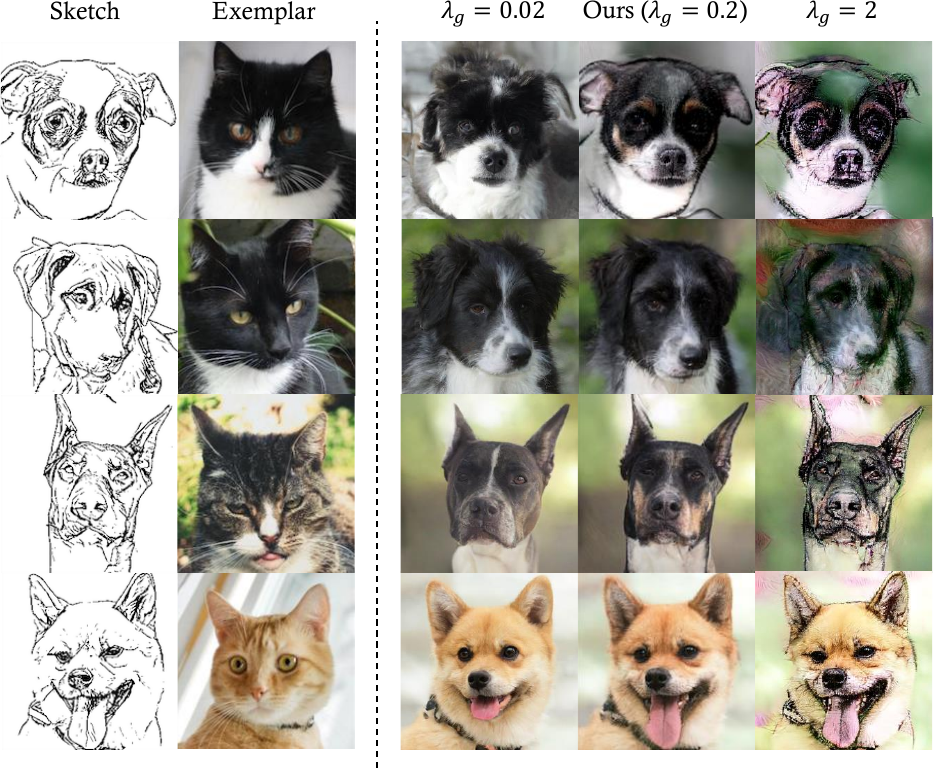}
\vspace{-1em}
\caption{
The results of different $\lambda_g$, and $\lambda_a=2$. With the increase of $\lambda_g$, the samples are more closer to sketch, resulting in lower realism.
}
\label{fig:shape_lam_ablation}
\end{figure}

\noindent\textbf{Effectiveness of the Shape-Energy Function.}\quad 
In Figure~\ref{fig:ablation_geom_metric}, we compare the results when using $L_2$ distance or $L_1$ distance as the similarity function in the shape-energy function $\mathcal{S}_g (\cdot,\cdot)$. We can observe that using $L_2$ distance can better preserve the shape of the input sketch. In the last column, we also provide an example when the shape-energy function is excluded. It turns out the shape of the input sketch is completely lost in the generated photo. The generation results demonstrate that the proposed shape-energy function with $L_2$ is a better choice for maintaining the geometric structure of input sketch. Additionally, in Figure~\ref{fig:shape_lam_ablation}, we illustrate the generated photos when using different $\lambda_g$. When the value of $\lambda_g$ is too large, the generated photos will become less realistic.

\begin{figure}[h]
\centering
\includegraphics[width=1\linewidth]{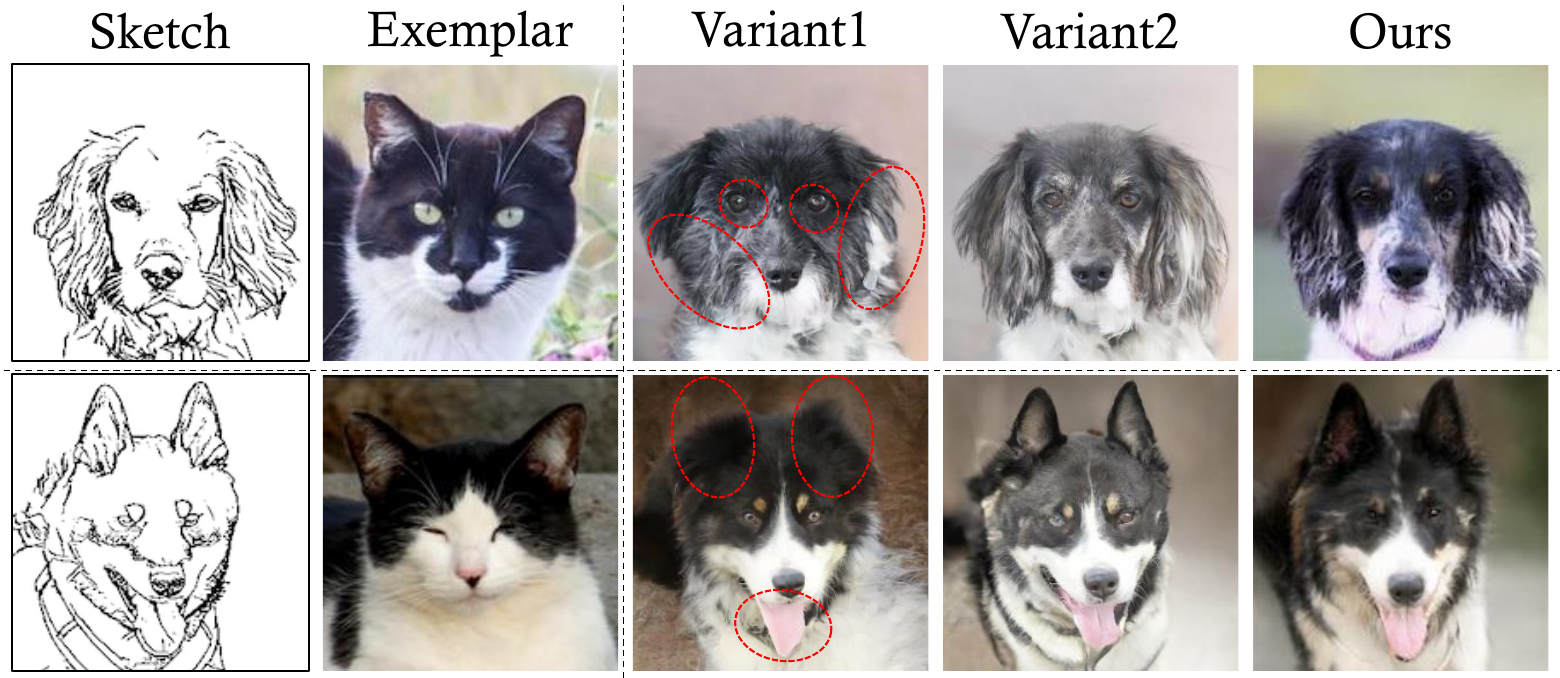}
\caption{
Results of our variant methods in terms of quality. Variant 1 directly takes the sketch as input and performs our full-control inversion. Variant 2 mixes up the input sketch and the exemplar, and then performs our full-control inversion.
}
\label{fig:compared}
\end{figure}

Furthermore, we compare with two model variants and show the results in Figure~\ref{fig:compared}. ``Variant 1'' directly takes the sketch as the input and performs full-control inversion. ``Variant 2 simply mixes up the sketch and the exemplar by calculating the mean value of these two images and then uses the combined image as the input to perform our full-control inversion. The results of Variant 1 show that the model confuses shape control with appearance control, resulting in generated photos losing their intended shapes. For example, the face and eyes of the dog in the first example is similar to the cat in the exemplar image. On the other hand, the colors and textures are not preserved in the generated photos of the Variant 2. These results suggest the necessity of our proposed shape-enhancing step.

\begin{table}[h]
\centering
\caption{
Quantitative analysis of $\Psi_i$ network. The $L_2$ distance represents the structural similarity between the sketch of the generated photos and the input sketches. PSNR evaluates the similarity between the generated photos and the exemplars.
}
\label{tab:feature_texture_compares}
\resizebox{\linewidth}{!}{
\begin{tabular}{l|cc|ccc}
\toprule
\multirow{2}{*}{$\Psi_i$ network} & \multicolumn{2}{c|}{Effect Control} & \multicolumn{3}{c}{Cat2Dog} \\
\cmidrule{2-6} & Shape & Texture & FID $\downarrow$ & $L_2$ $\downarrow$ & PSNR $\uparrow$ \\
\midrule
Without $\Psi_i$ & $\checkmark$ & $\checkmark$ & 41.21 & \textbf{28.17} & 5.49 \\
VGG~\cite{vgg_Liu_2015} & $\checkmark$ & $\checkmark$ & 43.40 & 30.13 & 5.49 \\
InceptionV3~\cite{inception_szegedy_2015} & $\checkmark$ & $\checkmark$ & 46.61 & 28.65 & 5.73 \\ 
CLIP~\cite{CLIP_Radford_2021} (Ours) & $\checkmark$ & $\checkmark$ & \textbf{39.84} & 28.22 & \textbf{5.89} \\ 
\bottomrule
\end{tabular}
}
\end{table}

\begin{figure}[h]
\centering
\includegraphics[width=\linewidth]{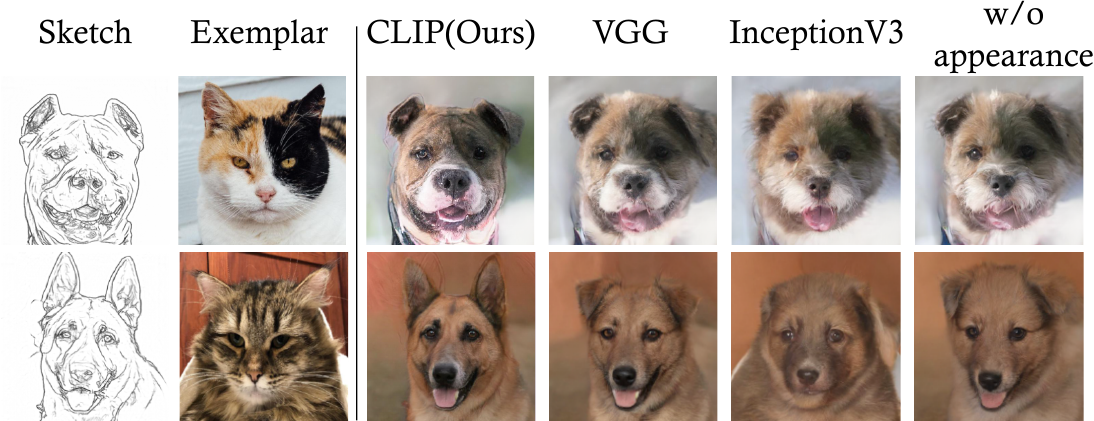}
\vspace{-1em}
\caption{The effectiveness of the proposed appearance-energy function.
}
\label{fig:feature_texture_netowrks}
\end{figure}
\noindent\textbf{Effectiveness of the Appearance-Energy Function.}\quad 
When attempting to perform image translation using full-control inversion, the absence of the appearance-energy function can lead to failure due to the sparse nature of the sketch, as depicted in Figure~\ref{fig:feature_texture_netowrks} "w/o appearance". This is because the diffusion model solely relies on the sketch as a constrain, without incorporating color information into the inversion process. As a result, the inversion process may fail. This proves the necessity of our proposed full-control inversion. 

Besides, choosing an appropriate neural network for the appearance-energy function is crucial for effectively learning appearance features. Therefore, we conducted an experiment to compare the performance of three networks, as illustrated in Figure~\ref{fig:feature_texture_netowrks}. We can observe that using CLIP~\cite{CLIP_Radford_2021} as the feature extractor yields the best results. Table~\ref{tab:feature_texture_compares} presents the quantitative results of different model variants, confirming the superiority of employing CLIP for the appearance-energy function.

\begin{figure}[t]
\centering
\includegraphics[width=1\linewidth]{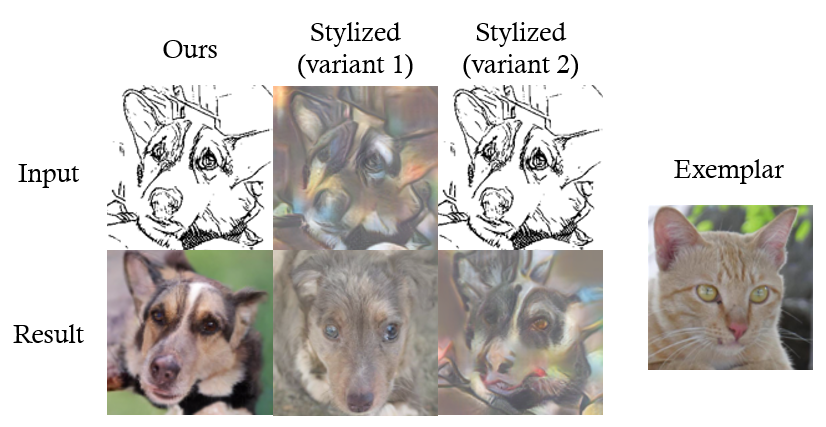}
\vspace{-1em}
\caption{
Using stylization for controlling the appearance. We provide the generation results of our proposed Inversion by Inversion method and other alternative methods that directly adopted the AdaIN~\cite{adain_xun_2017} to transfer the appearance of the exemplar instead of using our proposed appearance-energy function. ``Stylized (variant 1)" use AdaIN to add the appearance to the sketch before the inverse SDE process.  ``Stylized (variant 2)" uses AdaIN to add the appearance after the inverse SDE process.
}
\label{fig:different_stages}
\end{figure}
\noindent\textbf{Using Style Transfer to Control the Appearance.} \quad
It is an intuitive solution to use the style transfer method for this task. Figure~\ref{fig:different_stages} illustrates the generation results of simply using the style transfer method AdaIN~\cite{adain_xun_2017} to add the appearance of the exemplar. ``Stylized (variant 1)" use AdaIN to add the appearance to the sketch before the inverse SDE process while  ``Stylized (variant 2)" uses AdaIN to add the appearance after the inverse SDE process. It is observed that Stylized (variant 1), which performs the style transfer before the inverse SDE process, will significantly destroy the structure of the input sketch and thus the generation results is much worse than our method. Additionally, performing style transfer after the inverse SDE process (\textit{i.e.,} Stylized (variant 2)) fails to generate photo-realistic images, which also demonstrates the effectiveness of our shape-enhancing inversion step in preserving shape and the appearance-energy function in appearance control.

\section{Discussion}
\label{supp:analysis}

\noindent\textbf{Why does Inversion-by-Inversion work?}\quad It is challenging for a diffusion model to synthesize a photo based on a given sketch. A sketch only contains black and white pixels, making it hard for diffusion-based methods to generate photo-realistic images with different appearances, \textit{e.g.}, colors, and textures. Therefore, we turn to exemplar images to control appearance. However, existing diffusion-based models cannot handle two conditional images (sketch and exemplar) well, thus the quality of synthesized results is poor (as shown in Fig.~\ref{fig:regular_sketch_compare} and Fig.~\ref{fig:free_hands_sketch}).

The above issues motivate us to propose a new exemplar-based sketch-to-photo synthesis method called \textit{Inversion-by-Inversion}, which consists of two inversion SDE, a shape-enhancing inversion and a full-control inversion.
We believe that the two-stage inversion pipeline has two advantages over existing methods: \textit{First}, it naturally disentangles shape control and appearance control. Only the sketch is used as guidance for shape control during the first inversion process. While during the second inversion process, both sketch and exemplar are used, in which the exemplar dominates the guidance for appearance control. Such design significantly addresses the conflicts between these conditional images. Moreover, it progressively diminishes the domain gap between sketches and photos, improving the quality of the synthesis results.
\textit{Second}, the first inversion stage, i.e., shape-enhancing inversion, ensures the shape control from a sketch. As a sketch is sparse, its role could be shadowed by an exemplar image. Our design gives the sketch a higher priority than the exemplar through a separate inversion process. As a result, the sketch information can be enhanced.

\section{Conclusion}
In this work, we have proposed a new exemplar-based sketch-to-photo method called Inversion-by-Inversion based on effective stochastic differential equations (SDE). The model consists of two inversion stages, shape-enhancing inversion and full-control inversion. We have addressed the importance of the shape-enhancing step in enabling a diffusion-model-based method to handle sketch images. Besides, we have introduced a shape-energy function and an appearance-energy function for sketch and exemplar control, respectively. Through extensive experiments, we have shown the superiority of our proposed method in this task. Furthermore, our model can accept various types of exemplar images and synthesize high-quality photos from free-hand sketches.

\section{Acknowledgement}
This work is supported by the National Natural Science Foundation of China under Grant 62002012.

{\small
\bibliographystyle{ieee_fullname}
\bibliography{main}
}

\end{document}